%% file: main.tex
\documentclass[10pt,twocolumn,letterpaper]{article}

\usepackage[pagenumbers]{cvpr} 

\input{preamble}

\definecolor{cvprblue}{rgb}{0.21,0.49,0.74}
\usepackage[pagebackref,breaklinks,colorlinks,allcolors=cvprblue]{hyperref}

\def\paperID{13604} 
\def\confName{CVPR}
\def\confYear{2025}

\def\sysName{DejaVid\xspace}
\title{\sysName: Encoder-Agnostic Learned Temporal Matching for Video Classification}

\author{\textbf{Darryl Ho}\\
MIT CSAIL\\
Cambridge MA, USA\\
{\tt\small darrylho@csail.mit.edu}
\and
\textbf{Samuel Madden}\\
MIT CSAIL\\
Cambridge MA, USA\\
{\tt\small madden@csail.mit.edu}
}

\usepackage{pgf}
\newcommand{\myrounding}[1]{%
   \pgfmathparse{%
     round(#1*10)/10 - 0.0*0.1*((round((round(#1*100) - floor(round(#1*100) / 10) * 10)/10) - round((round(#1*100) - floor(round(#1*100) / 10) * 10 - 1)/10)) * (1-(floor(round(#1*100)/10) - floor(round(#1*100)/20)*2)))
   }%
   \pgfmathprintnumber[fixed, precision=1, zerofill, assume math mode=true]{\pgfmathresult}%
}

\usepackage{xcolor}
\definecolor{mygray}{gray}{0.5}  

\usepackage{multirow}

\def\vmaev{VideoMAE V}
\def\ssv{Something-Something V}
\def\kine{Kinetics-400\xspace}
\def\hmdb{HMDB51\xspace}

\def\ssvs{Sth-Sth2}
\def\kines{K400}
\def\hmdbs{HMDB}

\begin{document}
\maketitle
\input{sec/0_abstract}    
\input{sec/1_intro}
\input{sec/2_relatedwork}
\input{sec/3_method}

\input{sec/4_evaluation}
\input{sec/5_conclusion}
\input{sec/6_futurework}
\clearpage 
{
    \small
    \bibliographystyle{ieeenat_fullname}
    \bibliography{main}
}

\input{sec/X_suppl}

\end{document}

%% file: preamble.tex
\newcommand{\red}[1]{{\color{red}#1}}


%% file: sec/0_abstract.tex
\begin{abstract}

In recent years, large transformer-based video encoder models have greatly advanced state-of-the-art performance on video classification tasks. However, these large models typically process videos by averaging embedding outputs from multiple clips over time to produce fixed-length representations. This approach fails to account for a variety of time-related features, such as variable video durations, chronological order of events, and temporal variance in feature significance. While methods for temporal modeling do exist, they often require significant architectural changes and expensive retraining, making them impractical for off-the-shelf, fine-tuned large encoders. To overcome these limitations, we propose \sysName, an encoder-agnostic method that enhances model performance without the need for retraining or altering the architecture. Our framework converts a video into a variable-length temporal sequence of embeddings (TSE). A TSE naturally preserves temporal order and accommodates variable video durations. We then learn per-timestep, per-feature weights over the encoded TSE frames, allowing us to account for variations in feature importance over time. We introduce a new neural network architecture inspired by traditional time series alignment algorithms for this learning task. Our evaluation demonstrates that \sysName\ substantially improves the performance of a state-of-the-art large encoder, achieving leading Top-1 accuracy of 77.2\% on \ssv2, 89.1\% on \kine, and 88.6\% on \hmdb, while adding fewer than 1.8\%  additional learnable parameters and requiring less than 3 hours of training time. Our code is available at \url{https://github.com/darrylho/DejaVid}.
\end{abstract}

%% file: sec/1_intro.tex
\vspace{-.2in}
\section{Introduction}
\label{sec:intro}
\begin{figure}
\centering
\includegraphics[width=\linewidth]{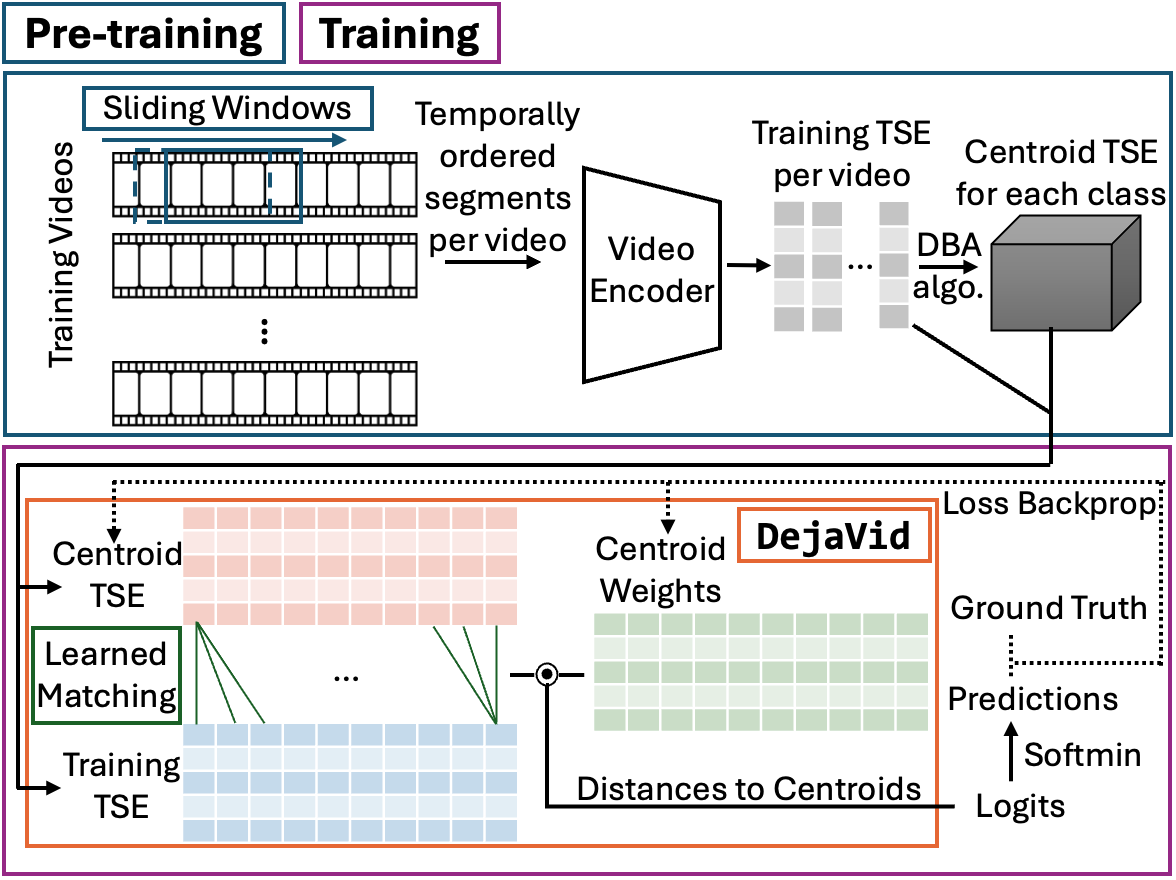}
\includegraphics[width=\linewidth]{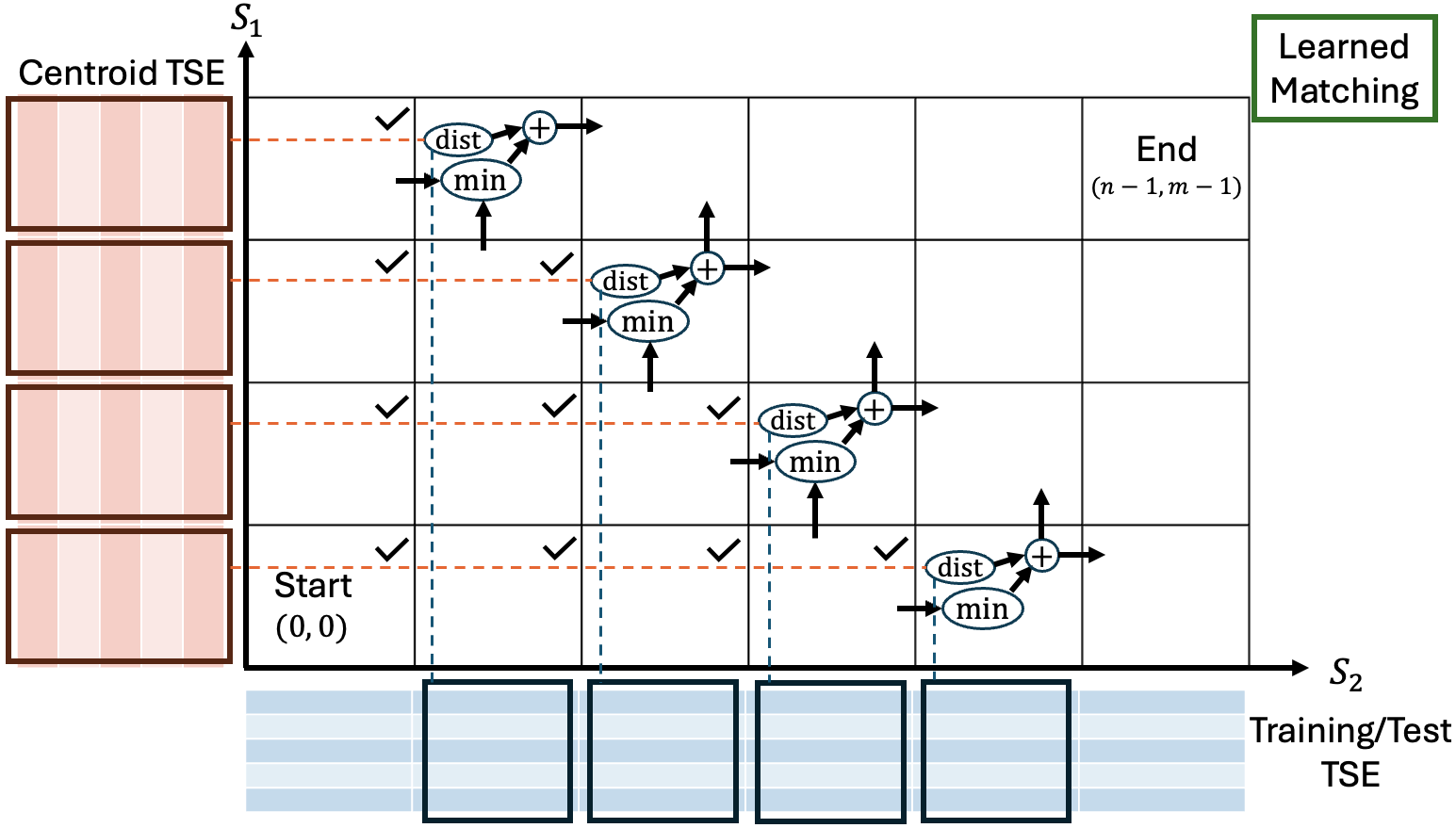}
\vspace{-.2in}\caption{{\bf \sysName system diagram.} We first transform a video into a variable-length temporal sequence of embeddings (TSE) then use Dynamic Time Warping Barycenter Averaging (DBA) \cite{dba} to calculate a reasonably ``average" TSE per class for its centroid initialization. We then learn per-timestep, per-feature weights for the encoded TSE frames, enabling us to accommodate variations in feature importance over time, as well as finetune the centroid TSE for each output class. For this task, we introduce a new neural network architecture inspired by traditional time series alignment algorithms. In testing, we freeze the post-training centroids and weights and use the same pre-training pipeline to generate TSEs to feed into \sysName for predictions.}
   \label{fig:sysdiag}
   \vspace{-.2in}
\end{figure}
\vspace{-.05in}Video action recognition has seen remarkable progress with the rise of transformer-based video models \cite{quovadisaction, vit, vivit}. To improve their performance, one of the most straightforward and effective approaches is to upscale the model and training data \cite{scalingvit}. As such,  the leaderboards of major action recognition benchmarks have come to be dominated by 100M--1B+ parameter large video transformers \cite{ssv12, kine, hmdb, vmaev2, iv2, mvdformultiviewfusion, tubevitformultiviewfusion}. 

These large encoders process videos with fixed frame counts, heights, and widths, generating a single embedding. On datasets with varying durations, the common approach is to pass multiple temporal clips and spatial crops through the encoder and then average the resulting logits for class prediction \cite{vmaev2, iv2, mvdformultiviewfusion, tubevitformultiviewfusion}. However, while the height and width of videos in a dataset are usually constant, their duration can vary significantly \cite{ssv12}. Furthermore, temporal order is far more critical. Unlike image flipping \cite{imagenet}, reversing time can alter semantics, such as opening vs. closing a door. When averaging logits from different temporal clips, sequential information in these clips is lost. Moreover, the significance of each feature within a video can change over time. For example, in a basketball three-point shot, early frames emphasize player position, while later frames focus on whether the ball goes in. Although encoders might not explicitly model semantically interpretable features like these, temporal variation in feature importance can also occur for arbitrary features, which mean-pooling fails to capture adequately. This motivates a key observation of our work: by failing to carefully account for temporal dimensions, state-of-the-art large models leave significant accuracy gains on the table.

Improving the temporal performance of large encoders is not trivial. Existing temporal-focused methods require invasive changes to the model, such as inserting temporal-specific layers between transformer blocks \cite{ila, atm23} or creating model duplications to focus on different input views \cite{mtv, slowfast}. These approaches are costly to apply to off-the-shelf pre-trained encoders because they require extensive changes in implementation and costly re-training. 

These insights motivate \sysName, a lightweight encoder-agnostic temporal approach that substantially boosts model classification performance. The simple yet crucial idea behind \sysName\ is that a video is more effectively represented as a temporal sequence of embeddings (TSE) rather than as a single embedding. This representation naturally preserves chronological order while supporting varying time lengths,  addressing the aforementioned shortcomings. \sysName\ is naturally able to use any pre-trained model by consuming a sliding-window of encoded video to generate a TSE, without requiring changes to the model’s architecture, retraining, or fine-tuning. \sysName\ is inspired by the well-known Dynamic Time Warping (DTW) algorithm \cite{dtw}, which dynamically aligns similar data points between two temporal sequences while maintaining their chronological order. In particular, \sysName\ introduces a novel neural network architecture inspired by DTW but with learned per-timestep, per-feature weights to handle the temporal variance of feature importance. Additionally, we develop a custom high-performance CUDA kernel for efficient computation and backpropagation of the (time-weighted) warping distance that outperforms a naive DTW-based implementation by 2 orders of magnitude.

We evaluate \sysName by layering it on top of the largest (ViT-g) variant of \vmaev2 \cite{vmaev2} (abbr. \vmaev2-g) as the backbone encoder to evaluate \sysName\ on three popular action recognition datasets: \kine\ \cite{kine}, \ssv2 \cite{ssv12}, and \hmdb\ \cite{hmdb}. We show that \sysName\ significantly improves on the strongest \vmaev2-g baseline, thereby achieving new state-of-the-art performance. In particular, \sysName\ achieves Top-1 accuracy of 77.2\% on \ssv2, 89.1\% on \kine, and 88.6\% on \hmdb, while adding fewer than 1.8\% additional learnable parameters and requiring less than 3 hours of training time. We also show that our proposed modifications to DTW, including the temporal weights, can further improve prediction performance. 

%% file: sec/2_relatedwork.tex
\vspace{-.05in}\section{Related Work}
\vspace{-.05in}\textbf{State-of-the-Art in Action Recognition.} Major action recognition benchmarks are topped by massive video transformers with hundreds of millions or billions of parameters. As of the time of writing, according to Papers with Code, the top performers on \kine\ and \ssv2 with verifiable parameter counts have between 600M and 6B parameters \cite{ssv12, kine, vmaev2, iv2, mvdformultiviewfusion, tubevitformultiviewfusion}. However, their evaluations on these two datasets all apply spatial and temporal mean-pooling to the logits at the end, which suffers from the time-related shortfalls related to variable video durations, chronological order of events, and temporal variance in feature significance, as described in \cref{sec:intro}.

\noindent\textbf{Temporal-focused Methods for Action Recognition.} There have been various works on improving the temporal modeling capability of video transformers. One general approach is to introduce temporal-specific layers in between transformer blocks. For example, ILA employs a mask-based layer that predicts interactive points between frame pairs to implicitly align features \cite{ila}, ATM applies basic arithmetic operations to pairs-of-frame features between ViT blocks \cite{atm23}, and TAM inserts an attention-based temporal adaptive module between 2D convolutions within a ResNet block \cite{tam21}. Another line of work is to fuse multiple temporal views in different ways. For example, SlowFast networks feed two different frame rates of a video into separate encoders and add lateral connections between them to fuse information across views \cite{slowfast}, and MTV deploys a similar idea but for both spatial and temporal dimensions \cite{mtv}. However, their performance is not competitive against the aforementioned large state-of-the-art models, partly due to their smaller size. Implementing these methods within the large models would involve considerable implementation changes and expensive retraining.

\noindent\textbf{Learning Temporal Sequence Alignments.} DTW is one of the most classic methods for calculating similarities between univariate time series or multivariate ones like our TSEs \cite{dtw, dtwreview}. Its traditional formulation is a dynamic programming scheme, but recent work has explored making it differentiable and then applying it to various fields. Soft-DTW \cite{softdtw} shows that DTW makes a good loss function when the min operator is replaced with soft-min. DTWnet \cite{dtwnet} proposes using DTW learning for time series feature extraction, and D3TW, OTAM, and FTCL \cite{d3tw, otam20, ftcl} similarly do this for weakly-supervised and few-shot video understanding. While these works have some similarities to our DTW-based temporal representation learning,  our method is technically different and superior in three ways: 
\begin{enumerate}
    \item \sysName models and learns the temporal variance in feature importance, which none of the other methods do. While classical DTW typically only works for a few dimensions (10-100) that require handcrafting or handpicking \cite{bagnall2018uea}, by adding learned weights, we extend DTW to high-dimensional use cases (1000-10000+) like video recognition by letting it learn the important dimensions.
    \item We reformulate our DTW-inspired temporal learning algorithm as a new neural network architecture that boosts parallelizablity and facilitates differentiability.
    \item We introduce changes to DTW to improve model stability, including removing the diagonal transition of DTW.
\end{enumerate}

\label{sec:relatedwork}

%% file: sec/3_method.tex
\section{Method}
\label{sec:method}

As described in \cref{sec:intro}, we represent a video as a temporal sequence of embeddings (TSE). In this section, we assume all videos have been converted into TSEs, and operate in this TSE space. \sysName\ feeds a video encoder a sliding-window stream of frames of an input video to generate a TSE, so the shape of a TSE is $T\times N_f$, where each index along $T$ represents the encoding of a window of frames, and thus $T$ can vary depending on individual video duration. $N_f$ is the size of the embedding vector of the backbone encoder and is constant.

Our problem is to classify TSEs. Specifically, we are given a training set of TSEs with class labels among $N_c$ classes, and our goal is to predict the classes of our validation TSEs. We  solve this problem in three broad steps:

\begin{enumerate}
    \item  For each class, we calculate a centroid TSE from the training TSEs of that class. 
    \item We optimize the centroid TSEs according to some loss function w.r.t. the training TSEs. 
    \item For each validation TSE, we calculate which class's centroid TSE is the most similar to it.
\end{enumerate}
    
\noindent Here, Steps 1 and 3 require a definition of distance (or similarity) between TSE pairs, and Step 2 requires the distance metric to be backpropagateable. The following \cref{subsec:def,subsec:backprop} describe how we solve these two problems. 

\subsection{Definitions}
\label{subsec:def}
There are multiple ways in which one can define the distance between two TSEs \cite{dtw,msm}, but in this paper, we focus on a standard time series distance metric: Dynamic Time Warping (DTW), as given in \cref{alg:dtw} (without the red text). DTW aligns two sequences that vary in time or speed by minimizing the distance between corresponding points. The two sequences are aligned by aligning each timestep in the first sequence to one or more consecutive timesteps in the second sequence while maintaining a monotonic relationship where a later timestep in the first must map to a later interval in the second. Monotonicity is maintained in Line 5 of \cref{alg:dtw}, where each step only moves forward in time for either sequence (from $i-1$ to $i$ or $j-1$ to $j$). This alignment minimizes cumulative distance and is commonly visualized as a ``warping path'' on a 2D grid, which shows the optimal alignment of the two sequences, each depicted in one grid axis. \begin{algorithm}
\caption{The Standard DTW Algorithm {\small{(without diagonal transition)}} \text{vs. the Time-Weighted DTW Algorithm \red{(red)}}}
Input: Two TSEs $a, b \in \mathbb{R}^{n \times N_f}, \mathbb{R}^{m \times N_f}$; a distance function $dist_{\red{w}}$\red{; $u\in \mathbb{R}_{>0}^{n \times N_f}$ temporal feature weights of $a$}

Output: Distance between $a$ and $b$
\begin{algorithmic}[1]
    \STATE $D \leftarrow \text{array}[n, m]$ \COMMENT{Out-of-bounds access returns $+\infty$}
    \STATE $D_{0,0} \leftarrow dist_{\red{w}}(\red{u_0,} a_0, b_0)$
    \FOR{$i$ in $[0, n-1]$} 
        \FOR{$j$ in ($[0, m-1]$ if $i>0$ else $[1, m-1]$)}
            \STATE $D_{i,j} \leftarrow \min(D_{i-1, j},D_{i, j-1}) + dist_{\red{w}}(\red{u_i,} a_i, b_j)$ 
        \ENDFOR 
    \ENDFOR 
    \RETURN $D_{n-1, m-1}$
\end{algorithmic}
\label{alg:dtw}\label{alg:wdtw}
\end{algorithm}\vspace{-.1in}

In the standard \cref{alg:dtw}, we use the Manhattan distance for the pointwise distance function $dist$. Also, although the original DTW algorithm includes a diagonal transition $D_{i-1,j-1}$ in the $\min$ clause, we omit this in \sysName because it improves model
stability. We discuss this decision in more detail in \cref{subsec:modperfopt}.

Let $D_v(a, b)$ denote the DTW distance between TSEs $a$ and $b$ as returned by the standard \cref{alg:dtw}, and let $C\in \mathbb{R}^{N_c\times T_c \times N_f}$ represent the centroid TSEs for the $N_c$ classes. Intuitively, we want to assign each TSE to the class it is nearest to in the DTW space. To do this, we define the predicted probability distribution $\hat{\mathbf{y}}_v$ for a training or validation TSE $m\in \mathbb{R}^{T_m \times N_f}$\footnotemark, \footnotetext{In practice, we perform a LayerNorm operation before the soft-min, but this doesn't qualitatively change our approach.}and the loss $\mathcal{L}_v$ for $m$ under centroids $C$ and one-hot ground truth vector $\mathbf{y}$:
\vspace{-.06in}\[\hat{\mathbf{y}}_v(C, m) := \underset{0\leq i < N_c}{\text{softmin}}(D_v(C_i, m))\]
\vspace{-.18in}\[\mathcal{L}_v(C, m) := \text{CrossEntropy}(\mathbf{y}, \hat{\mathbf{y}}_v(C, m))\]
However, as described in \cref{sec:intro}, we observe that the importance of each feature of a video can vary as time progresses. This observation motivates our time-weighted DTW distance metric, given in \cref{alg:wdtw} with modifications shown in red. The difference is the additional weight input $u$, which denotes the change in feature importance over time for $a$. We also modify the pointwise Manhattan distance function to incorporate $u$: $dist_w(u_i, a_i, b_j) := \sum_{k=0}^{N_f-1}u_{i,k}|a_{i,k}-b_{j,k}|$. The changes are colored in Lines 2 and 5 of \cref{alg:wdtw}. Intuitively, when \cref{alg:wdtw} calculates the distance between two sequences, our time-weighted modification allows it to weight some timesteps more than some other timesteps in the first series, and different features can have different weights as well. 
Let $D_w(u, a, b)$ denote the time-weighted DTW distance between TSEs $a$ and $b$ as returned by the time-weighted \cref{alg:wdtw}. We aim to learn the temporal variance of feature importance of the centroids $C$, so we define $U\in \mathbb{R}_{>0}^{N_c\times T_c \times N_f}$ of the same shape as $C$ to be the time-varying feature weights. To maintain the positivity of distances, we constrain $U$ to be positive. The time-weighted variant of predicted probability distribution $\hat{\mathbf{y}}_w$ and loss $\mathcal{L}_w$ are defined similarly as before\footnotemark[1]:
\vspace{-.06in}\[\hat{\mathbf{y}}_w(U, C, m) := \underset{0\leq i < N_c}{\text{softmin}}(D_w(U_i, C_i, m)) \]
\vspace{-.18in}\[\mathcal{L}_w(u, C, m) := \text{CrossEntropy}(\mathbf{y}, \hat{\mathbf{y}}_w(u, C, m))\]

\vspace{-.07in}From now on, we focus on the time-weighted \cref{alg:wdtw}, since one can set $U = \mathbf{1}_{N_c\times T_c \times N_f}$ to recover  standard DTW in \cref{alg:dtw}.

\vspace{-.05in}\subsection{Parallelizability \& Differentiability}
\label{subsec:backprop}
\begin{figure}[ht!b]\vspace{-.22in}
\centering
\includegraphics[width=\linewidth]{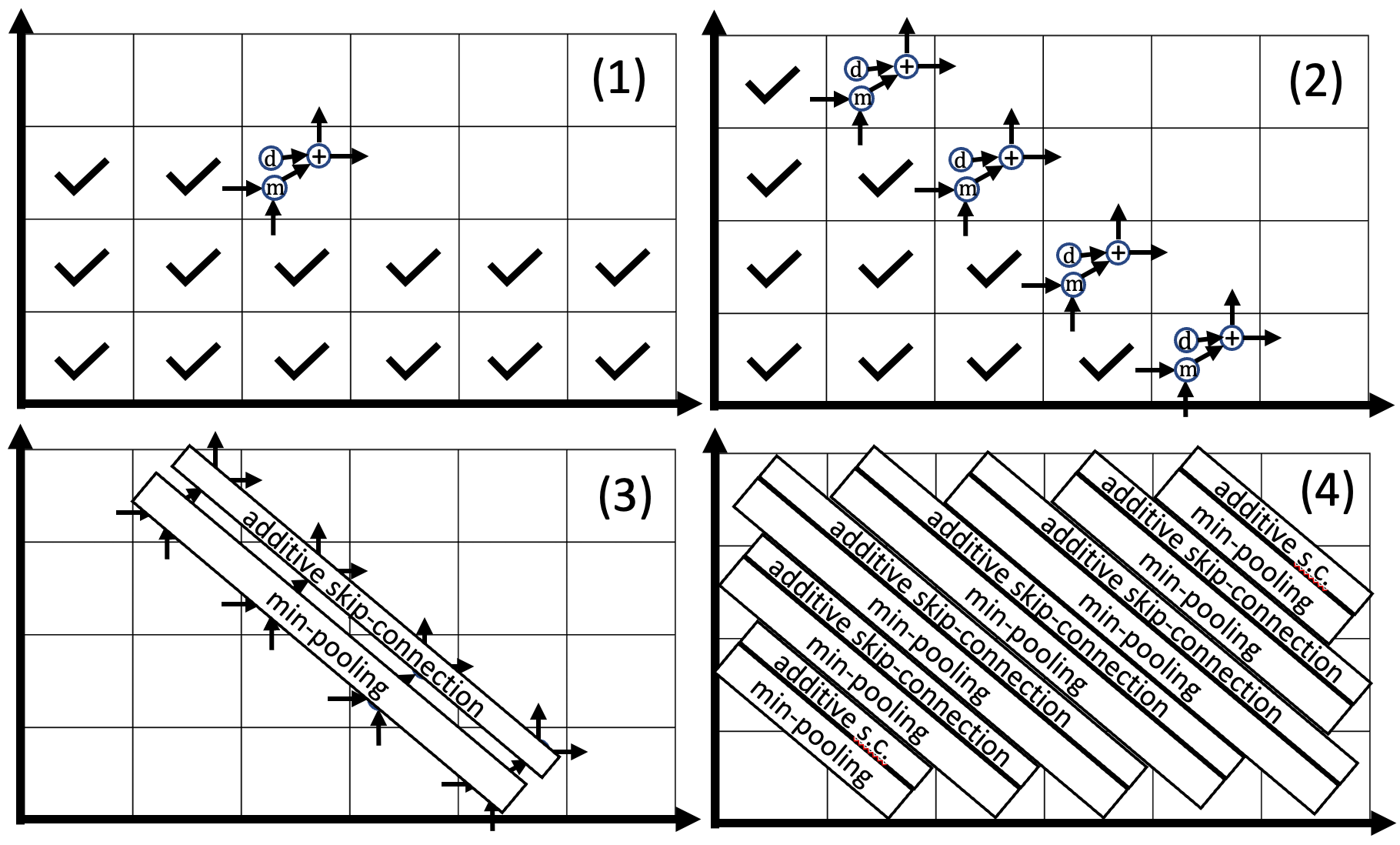}\vspace{-.15in}
\caption{1. Each cell of the DTW 2D grid executes a $min$ operation and then an $add\text{-}dist$ operation. Checkmarks show completed cells in row-major order. 2. In diagonal order, a diagonal can be executed as soon as the previous diagonal has finished. Checkmarks show completed cells in diagonal order. 3. We can view each diagonal as a min-pooling layer with kernel size 2 and stride 1, followed by an additive skip-connection layer. 4. We reformulate the algorithm into a serial stack of min-pooling layers interlaced with additive skip-connections. } 
   \label{fig:dtwnet}
   \vspace{-.1in}
\end{figure}

\cref{alg:nnwdtw} given an equivalent neural-network formulation for \cref{alg:wdtw};  this formulation facilitates the use of backpropagation for optimizing weights and enhances parallelizability. Before providing the details of the algorithm, we provide some intuition as to how and why it works.

\cref{fig:dtwnet} depicts the 2D grid commonly used for representing DTW warping path(s) between two sequences, as descrbed in \cref{subsec:def}. We visualize each operation in the nested loop in \cref{alg:wdtw} onto the corresponding cell in the grid. We can see that each cell executes a $\min$ operation on the outputs of its left-adjacent and bottom-adjacent cells, followed by an addition of some pointwise weighted distance $dist_w$. Note that cells are executed in a row-major order in \cref{alg:wdtw}. Therefore, the critical path length is quadratic ($O(nm)$). 

However, we leverage a critical property of \cref{alg:wdtw} that enables a reformulation of the algorithm, which is that \textbf{each cell can be executed as soon as its left and bottom neighbors have finished.} Therefore, we reformulate \cref{alg:wdtw} to execute in \textbf{diagonal order} and observe that \textbf{the $l$-th diagonal can be executed as soon as the $(l-1)$-th diagonal has finished.} Here, we define the $l$-th diagonal to be the cells $\{(i,j)\mid i+j=l\}$. For example, the top-right of \cref{fig:dtwnet} shows the 4th diagonal. Furthermore, elements in the same diagonal are \textbf{mutually independent}, so the work within each diagonal is parallelized. So we have reduced the critical path length to the number of diagonals, obtaining the linear runtime $O(n+m)$. 

Because each cell executes at Line 5 of \cref{alg:wdtw} a $\min$ operation on two consecutive cells of the previous diagonal and then adds some $dist_w$ term, we can express the new formulation as a neural net where each diagonal is a min-pooling layer with kernel size 2 and stride 1 followed by an additive skip-connection layer. Each min-pooling layer is executed after its previous skip-connection and min-pooling are completed. So \cref{alg:wdtw} is equivalent to a serial stack of min-pooling layers, with skip-connections interlacing in between. These descriptions are formalized in \cref{alg:nnwdtw}, where one can see the execution of the interlaced layers at Lines 10 and 11. For differentiability, we rewrite the time-weighted DTW distance $D_w$ as:
\vspace{-.1in}\[D_w:=SC_{n+m-2}+MP_{n+m-2}(SC_{n+m-3}+MP_{n+m-3}(\]
\vspace{-.282in}\[ \ldots + SC_1 + MP_1(SC_0) \ldots )) \]
where $MP_l$ and $SC_l$ are the $l$-th min-pooling and skip-connection as show in \cref{alg:nnwdtw}. Because both min-pooling and additive skip-connections are standard neural network operations, loss gradients trivially propagate through them. In a nutshell, the loss backpropagates to only the $dist_w(u_i, a_i, b_j)$ functions on the optimal warping path, which enables gradient-based optimization on $U$ and $C$. We leave the detailed formulas for loss gradients $\frac{\partial L_w}{\partial C}$ and $\frac{\partial L_w}{\partial U}$ in Supplementary Material.
\begin{algorithm}
\caption{Time-Weighted DTW Neural Network}
Input: Two TSEs $a, b \in \mathbb{R}^{n \times N_f}, \mathbb{R}^{m \times N_f}$; a distance function $dist_{w}$; $u\in \mathbb{R}_{>0}^{n \times N_f}$ temporal feature weights of $a$

Output: Distance between $a$ and $b$

\begin{algorithmic}[1]
    \STATE \COMMENT{Out-of-bounds access to $D$ or $SC$ returns $+\infty$}
    \STATE $SC_{0,0} \leftarrow dist_w(u_0, a_0, b_0)$ \COMMENT{The $0$-th skip-connection}
    \STATE $D_{0,0} \leftarrow SC_{0,0}$ 
    \FOR{$l$ in $[1, n+m-2]$}
        \STATE \COMMENT{The $l$-th diagonal}
        \STATE $\text{start}, \text{end} \leftarrow \max(0, l-m+1), \min(n-1, l)$
        \STATE $SC_{l} \leftarrow [dist_w(u_t, a_t, b_{l-t})]_{t=\text{start}}^{\text{end}}$
        \STATE $D_{l} \leftarrow \text{array}[\text{end}-\text{start}+1]$
        \STATE $\text{kernel}, \text{stride} \leftarrow 2, 1$
        \FOR{$i$ in $\text{range}(0, \text{end}-\text{start}+1, \text{stride})$}
            \STATE $D_{l,i} \leftarrow SC_{l, i} + \min_{s=0}^{\text{kernel}-1}(D_{l-1,i-s})$ 
        \ENDFOR 
    \ENDFOR 
    \RETURN $D_{n+m-2, 0}$

\end{algorithmic}
\label{alg:nnwdtw}
\end{algorithm}

\subsection{System Overview}
\label{subsec:sysoverview}
\vspace{-.05in}With the definitions introduced in \cref{subsec:def,subsec:backprop}, we now formalize the 3-step solution outlined at the start of \cref{sec:method}. \cref{fig:sysdiag} provides an overview of the \sysName system. The video encoder takes as input a video clip of fixed duration and outputs an embedding $e\in \mathbb{R}^{N_f}$. Given any off-the-shelf video encoder, we apply a sliding window over each video to produce a sequence of clips, which we then feed into the encoder to produce a temporal sequence of embeddings $E = \{e_0, e_1, \ldots, e_{T-1}\} \in \mathbb{R}^{T \times N_f}$, which forms a TSE for each video sample in the dataset. Then, we produce a centroid TSE of shape $T_c \times N_f$
\footnotemark\footnotetext{We set $T_c=8$ for all experiments in this work.} for each class by feeding the training TSEs to the DBA (Dynamic Time Warping Barycenter Averaging) \cite{dba} algorithm, which calculates a reasonable ``average" time series of the inputs while accounting for imperfect temporal alignment. Intuitively, each centroid TSE represents how the class-wide average of the $N_f$ features changes over time. Ultimately, we obtain many training TSEs, each with shape $T\times N_f$ for some varying $T$, and a stack of centroid TSEs $C$ of shape $N_c \times T_c \times N_f$, where $N_c$ is the number of classes. This concludes the pre-training stage of \sysName.  

In the training phase, to account for temporal variance in feature importance over time, each centroid TSE is paired with a positive weight tensor of the same shape $T_c \times N_f$ that represents the importance of each feature over time. The positivity is to maintain that the pointwise distance $dist_w$ is always positive. We denote the stack of weight TSEs $U$, with shape $N_c \times T_c \times N_f$. At initialization, the weight TSEs are either filled with ones or random positive values. 

In each epoch, we calculate the weighted DTW distance $D_w$ from each training TSE $m\in \mathbb{R}^{T \times N_f}$ to every centroid TSE to get the logit vector $\mathbf{z} = \{D_w(U_i, C_i, m)\}_{i=0}^{N_c-1} \in \mathbb{R}^{N_c}$, on which we  apply softmin to get the class prediction $\hat{\mathbf{y}}$ and cross-entropy loss $\mathcal{L}_w$ w.r.t. the ground truth $\mathbf{y}$. The weights $U$ and centroids $C$ TSEs are updated during backpropagation. To accelerate forward and backward passes, we implement a custom DTW CUDA kernel that efficiently calculates a slope tensor $S\in \mathbb{R}^{N_c \times T_c \times N_f}$ and an intercept tensor $B\in\mathbb{R}^{ N_c \times T_c \times N_f}$ such that $\texttt{torch.sum}(U\odot (C \odot S + B), \texttt{dim=(-2,-1)})$ equals the logit vector $\mathbf{z}$ and the effective gradients w.r.t. $U$ and $C$ equals $\frac{\partial L_w}{\partial U}$ and $\frac{\partial L_w}{\partial C}$. The implementation of the CUDA kernel, roughly speaking, leverages the fact that a weighted DTW distance is essentially the sum of many terms of $u\cdot|c-b|$, which equals $u\cdot(c\cdot(\pm 1) + (\mp 1)b)$, where the $(\pm 1)$ term contributes to slope $S$ and $(\mp 1)b$ to intercept $B$.

\vspace{-.05in}\subsection{Modifications to Standard DTW}
\label{subsec:modperfopt}
\vspace{-.05in}As mentioned in \cref{subsec:def}, unlike the original DTW algorithm, we exclude the diagonal transitions (i.e., transitioning from $D_{i-1, j-1}$ to $D_{i,j}$ in \cref{alg:dtw,alg:wdtw}) for \sysName. The reason is due to model stability. When matching two series of length $n$ and $m$, by disallowing diagonal transitions, the warping path from $D_{0,0}$ to $D_{n-1,m-1}$ always takes $n+m-1$ steps, and so the DTW distance will always be the sum of $n+m-1$ step-wise distances. If diagonal transitions are allowed, the number of step-wise distances on the warping path can vary, destabilizing the model. Although beyond the scope of this paper, we note that there could be other ways to deal with diagonal transitions, such as assigning or learning a multiplicative or additive distance bias for diagonal vs. non-diagonal transitions, such as  in Move-Split-Merge (MSM) \cite{msm}, a more modern replacement of DTW that treats diagonal transitions differently from non-diagonal ones.

Another modification we made is to fix epoch centroids and weights.  That is, at the beginning of an epoch, we make a frozen copy of the current centroids and weights, and then during the epoch, use them to calculate the warping path for each centroid-training TSE pair. Then, along the path, we calculate the pointwise distance using the (non-frozen) originals and perform backpropagation on them. Note that the frozen copies are not updated per batch but instead per epoch. The reason for this modification is again due to model stability. We found that if the warping path calculation mechanism is updated every batch, the distances become noisier from batch to batch, and the model has a harder time converging. 

We discuss the effect of these modifications in \cref{subsec:ablation}.

%% file: sec/4_evaluation.tex
\vspace{-.12in}\section{Evaluation}
\label{sec:evaluation}
\vspace{-.05in}In this section, we describe our implementation and evaluation of \sysName.
\vspace{-.05in}\subsection{Datasets and Evaluation Metrics}
\label{subsec:dataevalmet}
\vspace{-.05in}We test \sysName\ on three different popular action recognition datasets: \ssv2 \cite{ssv12}, \kine\ \cite{kine}, and \hmdb\ \cite{hmdb}. The first two are large-scale datasets: \ssv2 has 168,913 videos for training and 24,777 for validation, while \kine\ has 240,436 training and 19,787 validation videos. In contrast, \hmdb\ only has 3,570 training and 1,530 validation videos, but its smaller size can help us study the model's performance when the training set is smaller.

\ssv2 has 174 action classes that are object and motion-centric, which requires learning the subtle differences between motions and object interactions. \kine\ covers 400 classes of everyday human actions derived from real-world scenarios on YouTube and can be human-object or human-human interactions. Its videos are also longer than those of \ssv2 on average ($\sim{300}$ frames vs. $\sim 40$ frames), which can help differentiate the temporal modeling capabilities between models. \hmdb\ has 51 human-dominated action classes, like \kine, containing some individual and some human-human behaviors. 

For all these datasets, we train \sysName\ on the training sets and test on the validation sets using the same evaluation protocols as those in \vmaev2 \cite{vmaev2}. We report the top-1 and top-5 accuracy for every experiment on \ssv2 and \kine. Due to the smaller action set of \hmdb, we only report top-1 for experiments on it.  

\vspace{-.05in}\subsection{Implementation Details}\label{subsec:impldet}
\vspace{-.05in}\textbf{Pre-training.} We use the largest (ViT-g) model from \vmaev2 \cite{vmaev2} (abbr. \vmaev2-g) as the video encoder.\footnotemark\footnotetext{\vmaev2 publishes different finetuned weights for \kine and \ssv2, and the script for finetuning on \hmdb, which we can afford to run thanks to the small dataset size. We use the respective finetuned weights for the pre-training stage of each dataset.} \vmaev2-g is the only model we experiment with because it is the only SOTA model\footnotemark[4] that publishes finetuned weights on large datasets like \kine and \ssv2, which we cannot afford to finetune ourselves. To show that \sysName is model-agnostic, we apply the same method to other non-SOTA video encoders in Supplementary Material. The native embedding output length of \vmaev2 is $1408$, which is then converted into $N_c$ logits for class prediction, so we let $N_f:=1408+N_c$ denote the length of the entire output of \vmaev2-g. For \sysName, we apply a temporal sliding window across the input video $T$ times to feed into the encoder, resulting in a TSE of shape $T\times N_f$. We then randomly sample some TSEs for each action class to calculate its initial centroid TSE using the DBA algorithm \cite{dba}. $T$ can be different across different datasets and within the same dataset. In particular, we use $T=33$ for \kine and \hmdb, and $T=4|vid|$ for \ssv2, where $|vid|$ is the length of an individual video. For a detailed description of our pre-training configuration and the choice of width and stride for the sliding window of each dataset, please refer to the Additional Implementation Details section in Supplementary Material.


\vspace{.05in}\noindent \textbf{Training \& Testing.} The pre-training stage gives us a TSE $m$ for each training or validation video as well as the centroid TSEs $C$. The weight tensor $U$ is either randomly initialized or one-initialized (i.e., filled with ones). To enforce positivity, we actually store the log of the weights and exponentiate them during forward passes. These log-of-weights are randomly initialized or zero-initialized (i.e., filled with zeros). We use the AdamW optimizer with betas $(0.9, 0.999)$ and no weight decay. The initial learning rate for the log-of-weights is 1e-3 for both \ssv2 and \kine, and 8e-3 for \hmdb. The learning rates for the centroid TSEs are one-third of the log-of-weights. We use batch size  48 and train for 36 epochs. We use a cosine scheduler across all 36 epochs. A validation run is performed after every epoch, and we report the top accuracy recorded across all epochs.

\vspace{-.05in}\subsection{Main Results}\label{subsec:mainres}
\vspace{-.05in}This section compares our results with previous other temporal-related and state-of-the-art methods. The main baseline is \vmaev2-g, which is a billion-parameter video encoder that is state-of-the-art or near state-of-the-art on all evaluation datasets \cite{vmaev2, ssv12, kine, hmdb}. We test our method on each dataset with three different settings: random-initialization for the weight TSEs $U$ (denoted as \textit{full learning; w-init-rnd} in the following charts), one-initialization for $U$ (denoted as \textit{full learning; w-init-one}), and finally one-initialization for $U$ but freezing it throughout (i.e., disallowing \texttt{optimizer.step()} to act on $U$; denoted as \textit{frozen weights; w-init-one}). Note that the last setting is for experimenting with the unweighted DTW distance as in \cref{alg:dtw}, so frozen weights with random initialization don't make sense.

\vspace{.05in}\noindent\textbf{Results on \kine}. 
As shown in \cref{tab:kine}, our method achieves leading accuracy on \kine. We improve on the strongest \vmaev2-g baseline by 0.7\%.  \sysName\ only requires 5.8M parameters, less than 0.6\% of the encoder's parameter count, to achieve this gain. For comparison, the gain from \vmaev1 to \vmaev2 uses 380M extra parameters to increase the accuracy by only 0.4\%. Note that our results are achieved without re-finetuning the encoder, so the training cost is minimal compared to training the encoder. In short, \sysName\ is a comparatively tiny add-on that significantly boosts the performance of an off-the-shelf frozen model.

The performance gain of \sysName\ is not only due to temporal supersampling. We test this by averaging the temporal dimension of each sample TSE to get a length-$(1408+N_c)$ vector and using the last $N_c$ elements for class prediction; the result is shown as \vmaev2 (temporal supersampling) \cref{tab:kine}. When compared against the original setting at \vmaev2 $5\times 3$, we see this alone does not improve the accuracy. Also, the temporal weight-learning does not provide additional benefits on top of our centroid-finetuning for \kine. We suspect the reason is that the temporal weights increase the expressivity of our method, so overfitting becomes more likely; one future work is to look into regularization for \sysName.
\footnotetext[4]{We were unable to run InternVideo2 on our machine due to one of its required packages, \texttt{flash-attn}, being incompatible with our machine.}
\footnotetext[5]{The authors of \vmaev2 report 77.0 and 95.9 for top-1 and top-5 accuracy, respectively, but we measure only 76.7 and 95.8 using their published weights. We suspect the discrepancy is due to the input video loading mechanism. The official GitHub repo of \vmaev2 has inconsistent descriptions about the data preparation for \ssv2, but as of the time of writing, the source code assumes each video has been converted to a folder of jpeg images, and \vmaev2 loads these jpeg files instead of the source video. But when we do exactly that, the top-1 accuracy measures only 75.7. This is understandable since JPEG uses lossy compression. So we modified the source code to use opencv2 to directly load each video, which gives 76.7 and 95.8.}
\begin{table*}[t]
  \centering
  \resizebox{\textwidth}{!}{ 
      \begin{tabular}{@{}l|c|c|c|cc|c|c|cc|c|c|c@{}}
        \toprule
        \multirow{2}{*}{\centering Method} & \multirow{2}{*}{\centering W.A.} &  \multicolumn{4}{|c}{\kine} & \multicolumn{4}{|c}{\ssv2}& \multicolumn{3}{|c}{\hmdb}  \\
        & & Params & Clips×Crops& Top-1 & Top-5& Params & Clips×Crops& Top-1 & Top-5 &Params & Clips×Crops& Top-1\\
        \midrule
        \multicolumn{13}{c}{\color{mygray}Other State-of-the-Art Large Models} \\ 
        \color{mygray}TubeViT \cite{tubevitformultiviewfusion} &\color{mygray}No& \color{mygray}632M & \color{mygray}4×3 & \color{mygray}90.9 & \color{mygray}98.9 & \color{mygray}307M& \color{mygray}-& \color{mygray}76.1& \color{mygray}95.2&-&-&-\\
        \color{mygray}MVD \cite{mvdformultiviewfusion} &\color{mygray}No& \color{mygray}633M & \color{mygray}5×3 & \color{mygray}87.2 & \color{mygray}97.4& \color{mygray}633M & \color{mygray}2×3& \color{mygray}77.3& \color{mygray}95.7&-&-&-\\
        \color{mygray}InternVideo2-1B \cite{iv2} &\color{mygray}Yes\footnotemark& \color{mygray}1B & \color{mygray}4×3& \color{mygray}91.6& \color{mygray}-& \color{mygray}1B & \color{mygray}2×3& \color{mygray}77.1& \color{mygray}-&-&-&-\\
        \color{mygray}InternVideo2-6B \cite{iv2} &\color{mygray}No& \color{mygray}6B & \color{mygray}4×3& \color{mygray}92.1& \color{mygray}-& \color{mygray}6B & \color{mygray}5×3& \color{mygray}77.5& \color{mygray}-&-&-&-\\
        \midrule
        \multicolumn{13}{c}{Other Temporal-Focused Methods} \\ 
        ILA \cite{ila} &-& - & 4×3 & 88.7 & 97.8 & - & 4×3 & 70.2& 91.8&-&-&-\\
        ATM \cite{atm23} &-& - & 4×3 & \textbf{89.4} & \textbf{98.3}& - & 2×3& 74.6& 94.4&-&-&-\\
        TAM \cite{tam21} &-& - & 4×3& 79.3& 94.1& - & 2×3& 66.0& 90.1&-&-&-\\
        SlowFast \cite{slowfast} &-& - & 10×3& 79.8& 93.9& - & -& 61.7& -&-&-&-\\
        MTV \cite{mtv} &-& - & 4×3& \textbf{89.1}& \textbf{98.2}& - & -& 68.5& 90.4&-&-&-\\
        \midrule
                \multicolumn{13}{c}{Main Baselines} \\ 
 \vmaev1-H \cite{vmaev2} &Yes& 633M& 5×3 & 88.1&-&-&-&-&-&-&-&-\\
 
    {\vmaev1-L (32-frame 
 input) \cite{vmaev1,vmaev2}}&Yes& -&-&-&-&305M& 2×3& 75.4&95.2&-&-&-\\
        \vmaev2-g \cite{vmaev2} &Yes& 1013M & 5×3 &  88.4 & 98.0 & 1013M & 2×3&  76.7\footnotemark & 95.8\footnotemark[5]& 1013M & 5×3 & 88.1\\
        \vmaev2-g (temporal supersampling) \cite{vmaev2} &Yes& 1013M & 33\footnotemark×1& \myrounding{88.10} & \myrounding{97.77}& 1013M & $(4|vid|)$\footnotemark[6]×1& \myrounding{76.20} & \myrounding{95.69} & 1013M & 33\footnotemark[6]×1& \myrounding{88.04}\\
        
        \midrule
                \multicolumn{13}{c}{Our Results} \\ 
        Ours (frozen weights; w-init-one) &-& +5.8M &  & \textbf{\myrounding{89.08}}  & \textbf{\myrounding{98.19}} & +8.8M &  & \myrounding{77.05} & \textbf{\myrounding{96.26}}& +0.60M &  & \myrounding{88.30} \\
        Ours (full learning; w-init-rnd) &-& +11.6M & 33\footnotemark[6]×1&  \myrounding{88.86} &  \myrounding{98.06} & +17.6M & $(4|vid|)$\footnotemark[6]×1&  \textbf{\myrounding{77.16}} & \textbf{\myrounding{96.32}} & +1.19M & 33\footnotemark[6]×1&  \textbf{\myrounding{88.56}} \\
        Ours (full learning; w-init-one) &-& +11.6M &  &  \myrounding{88.99} &  \textbf{\myrounding{98.16}} & +17.6M &  &  \myrounding{77.04} & \textbf{\myrounding{96.31}}& +1.19M &  &  \textbf{\myrounding{88.56}}  \\
        \bottomrule
      \end{tabular}
  }
  
  \caption{Comparison with the \vmaev2-g baseline and other relevant variants, SOTA, or temporal methods on \kine, \ssv2, and \hmdb. All accuracies are reported in percent (\%). W.A. = Weights Available. Dashes (-) = N/A.}
  \vspace{-.25in}
  \label{tab:kine}\label{tab:ssv2}\label{tab:hmdb}
\end{table*}
\footnotetext{The definition of these clips are described in \cref{subsec:impldet}.}

\vspace{.05in}\noindent\textbf{Results on \hmdb.} The results in \cref{tab:hmdb} show that our model also achieves state-of-the-art accuracy on the \hmdb. We achieve a 0.5\% performance gain against the strongest \vmaev2-g baseline. To the best of our knowledge and according to Papers with Code, this is the highest accuracy recorded on the \hmdb\ action recognition benchmark to date (as of Nov, 2024). This shows that \sysName\ works well in a training data-scarce scenario. We draw similar conclusions to those from \kine\ in parameter efficiency -- our parameter count is 1.19M, which is less than 0.12\% of that of \vmaev2-g -- and temporal supersampling alone does not improve performance. 

\vspace{.05in}\noindent\textbf{Results on \ssv2.}  \cref{tab:ssv2} compares our results on \ssv2\ against other works. We outperform the top \vmaev2-g baseline with a performance gain of 0.5\%. To the best of our knowledge and according to Papers with Code, \sysName\ achieves the highest top-5 accuracy and one of the highest top-1 accuracies on the \ssv2 action recognition benchmark to date (as of Nov, 2024). 

Progress on \ssv2 has slowed in recent years, with top accuracy improving by no more than 0.3\% between 2022 and 2024 \cite{_pwcssv2}. Such gains often require massive model scaling; for example, InternVideo2 added over 5 billion parameters for a 0.4\% boost \cite{iv2}. In contrast, \sysName\ achieves comparable improvements with just 17.6M parameters—fewer than 1.8\% of \vmaev2-g. This efficiency highlights the significance of our +0.5\% accuracy gain, which equates to more than two years of recent progress \cite{mvdformultiviewfusion,iv2,vmaev2}.


However, the performance gain is smaller than on \kine. One possible reason could be the difference in average video length: $\sim 40$ for \ssv2 and $\sim 300$ for \kine. Because \kine videos are longer, there may be more temporal information for \sysName\ to capture. Another reason may be that the encoder is trained with variable video durations, as described in \cref{subsec:impldet}, so one future work would be to re-finetune the encoder with video clips of constant frame-count and retry our experiments. 

\vspace{-.05in}\subsection{Ablation Studies}
\label{subsec:ablation}
\vspace{-.05in}\noindent\textbf{Centroid vs. Weight Learning.} In \cref{tab:mwlearning}, we analyze the impacts of centroid and weight learning of \sysName\ by freezing one component at a time and comparing the results. As seen, both frozen-centroid and frozen-weight variants yield meaningful accuracy gains against the \vmaev2-g baseline, and so both components are indeed beneficial. The centroid component has a greater performance gain on average across the three datasets, giving +0.43\% top-1 accuracy gain vs. 0.25\% for the weights, but the weight component seems to be more useful when training data is scarce. As for the training data-rich cases, the weight component seems to be more useful in \ssv2 than in \kine, potentially because we have 4 different feature sets from the different sliding window widths, and these features should naturally have different importance, so the one-initialized weights are further away from the global optimal. 
\begin{table}[H]
  \centering
    \resizebox{\columnwidth}{!}{ 
  \begin{tabular}{@{}l|cc|cc|c@{}}
    \toprule
    Method & \multicolumn{2}{c|}{\ssvs} & \multicolumn{2}{c|}{\kines} & \hmdbs \\
     & top-1 & top-5 & top-1 & top-5 & top-1 \\
    \midrule
    \vmaev2-g @ best \cite{vmaev2} & 76.7\footnotemark[5]&  95.8\footnotemark[5]&  88.4&  98.0&  88.1\\ 
    \midrule
    Frozen centroids (w-init-rnd)& \myrounding{76.84}&  \myrounding{96.18}&  \myrounding{88.64}&  \myrounding{97.98}&  \myrounding{88.43}\\ 
    Frozen centroids (w-init-one)&  \myrounding{76.87}&  \myrounding{96.20}& \myrounding{88.75}&  \myrounding{98.07}&  \myrounding{88.43}\\ 
    Frozen weights (w-init-one)&  \myrounding{77.05} & \textbf{\myrounding{96.26}} & \textbf{\myrounding{89.08}} & \textbf{\myrounding{98.19}}  &  \myrounding{88.30}\\ 
    Full learning (w-init-rnd)& \textbf{\myrounding{77.16}} & \textbf{\myrounding{96.32}} &  \myrounding{88.86}&  \myrounding{98.06}&  \textbf{\myrounding{88.56}}\\
    Full learning (w-init-one)& \myrounding{77.04} & \textbf{\myrounding{96.31}}  & \myrounding{88.99} &  \textbf{\myrounding{98.16}}&  \textbf{\myrounding{88.56}}\\
    \bottomrule
  \end{tabular}
  }
  \vspace{-.1in}
  \caption{Ablation study on centroid \& weight learning.}
  \vspace{-.25in}
  \label{tab:mwlearning}
\end{table}
\begin{table}[H]
  \centering
  \resizebox{\columnwidth}{!}{
  \begin{tabular}{@{}l|cc|cc|c@{}}
    \toprule
    Method & \multicolumn{2}{c|}{\ssvs} & \multicolumn{2}{c|}{\kines} & \hmdbs \\
     & top-1 & top-5 & top-1 & top-5 & top-1 \\
    \midrule
    DT on, otherwise best &  \myrounding{77.02}&  \myrounding{96.23}&  \myrounding{88.57}&  \myrounding{98.00}&  \myrounding{(88.43+88.56)/2}\\ 
    FECW off, otherwise best&  \myrounding{77.10}&  \myrounding{96.19}&  \myrounding{89.03}&  \textbf{\myrounding{98.18}}&  \myrounding{(88.50+88.43)/2}\\ 
    Best (DT off, FECW on) &  \textbf{\myrounding{77.16}} &  \textbf{\myrounding{96.32}} &  \textbf{\myrounding{89.08}} &  \textbf{\myrounding{98.19}}  &  \textbf{\myrounding{88.56}}\\
    \bottomrule
  \end{tabular}
  }
  \vspace{-.1in}
  \caption{Ablation studies on diagonal transition (DT) and fixed epoch centroids \& weights (FECW).}
  \vspace{-.25in}
  \label{tab:dt} \label{tab:fecw}
\end{table}
\begin{table}[H]
  \centering
    \resizebox{\columnwidth}{!}{ 

  \begin{tabular}{@{}l|ccc@{}}
    \toprule
    Method& \ssvs & \kines & \hmdbs \\
    \midrule
    Naive DTW implementation& 4.79e4\footnotemark& 1.17e5\footnotemark[7]& 294\\ 
    Custom DTW CUDA kernel& \textbf{297}& \textbf{156}& \textbf{1.01}\\
    \bottomrule
  \end{tabular}
  }
  \vspace{-.1in}
  \caption{Ablation study on the custom DTW CUDA kernel. Table shows runtime per training epoch in seconds.}
  \vspace{-.15in}
  \label{tab:cudaperf}
\end{table}\footnotetext{Estimated by running 3 batches.}

\noindent\textbf{Diagonal Transition of DTW.} \cref{tab:dt} presents a comparison between the best (highest top-1 accuracy\footnotemark\footnotetext{In the case of tied top-1 accuracies, we apply the same change to each tied setting and average their resulting accuracies after the change.}) setting of \cref{tab:mwlearning} for each dataset and the same setting but with diagonal transitions turned on.  Enabling diagonal transitions consistently performs worse on all datasets. This justifies our choice of disabling diagonal transitions in \sysName\ and verifies the insight behind this choice as described in \cref{subsec:modperfopt}. 

\vspace{-.00in}\noindent\textbf{Fixed Epoch Centroids \& Weights for DTW Learning.} We use \cref{tab:fecw} to compare the best setting (highest top-1 accuracy\footnotemark[8]) from \cref{tab:mwlearning} for each dataset with the same setting but disabling fixed epoch centroids \& weights, where we do not make frozen copies of the centroids and weights for warping-path determination (shown as FECW off). Again, the inferior performance of this group compared to when FECW is on across all datasets justifies our design for \sysName, aligning with the reasons detailed in \cref{subsec:modperfopt} about model stability. However, the gap is less significant than that for the diagonal transition. This suggests there may be room for improvement by tweaking the path-determination mechanism, such as only updating every other epoch instead of each epoch.

\vspace{-.00in}\noindent\textbf{Custom CUDA Kernel for Accelerating DTW.}  \cref{tab:cudaperf} summarizes the runtime per epoch of the best-performing settings on all datasets compared to using a naive Python implementation of DTW.  All experiments are conducted on a single node with 8 NVIDIA H100 GPUs. We can see that our custom CUDA kernel provides a speedup between 161x and 750x. The most technically significant speedup comes from reducing the critical path length from quadratic to linear, as discussed in \cref{subsec:backprop}. Other factors contributing to the speedup include the Python/C++ speed difference and architecture and locality-aware, LibTorch-native implementation of the CUDA-kernel.

%% file: sec/5_conclusion.tex
\vspace{-.11in}\section{Conclusion}\label{sec:conclusion}
\vspace{-.08in}We presented \sysName, an encoder-agnostic method that boosts video classification performance without retraining the encoder or changing its architecture. Instead of converting a video into a single fixed-length embedding like most state-of-the-art encoders do, we feed a backbone encoder a sliding-window stream of frames to obtain a variable-length temporal sequence of embeddings (TSE), which preserves temporal order and supports variable video durations. We then introduce a novel neural network architecture based on Dynamic Time Warping (DTW) with learnable, per-timestamp feature weights to address temporal variance in feature significance. Our ablation studies compare and justify various design choices and modifications to classical DTW, and our evaluation shows that \sysName\ achieves state-of-the-art video classification results on a variety of benchmarks, significantly improves the accuracy of a cutting-edge large encoder, and outperforms other temporal methods. In particular, we achieve leading Top-1 accuracy of 77.2\% on \ssv2, 89.1\% on \kine, and 88.6\% on \hmdb, representing a 0.5-0.7\% accuracy gain over \vmaev2-g while adding less than 1.8\%  additional learnable parameters and less than 3 hours of training time.


%% file: sec/6_futurework.tex
\vspace{-.05in}\section{Future Work}\label{sec:futurework}
\vspace{-.05in}In this paper, we selected video classification as the evaluation task due to its foundational role in computer vision \cite{karpathy2014large,TranICCV15,kine}, its real-world importance \cite{AbuElHaija2016YouTube8M,slowfast,Girdhar2019VAT}, and the significant amount of highly competitive research focused on it \cite{vivit,vmaev1,vmaev2,tubevitformultiviewfusion,mvdformultiviewfusion,iv2,ila,atm23,tam21,mtv,slowfast}. With that said, we can also see the potential of \sysName\ -- and of the power of fine-grained temporal alignment learning in general -- in action segmentation, video captioning, text-to-video retrieval, and any other tasks that can benefit from detailed temporal modeling.

%% file: sec/X_suppl.tex
\clearpage
\setcounter{page}{1}
\maketitlesupplementary

{\noindent\bf Additional Implementation Details.}
\vmaev2-g takes 16 frames of shape $224\times 224$ as input and outputs a length-1408 representation and then a length-$N_c$ logit vector, where $N_c$ is the number of classes in the dataset. Thus the size of the encoder output $N_f$ is $1408+N_c$. For \sysName, we apply a temporal sliding window across the input video, take the center crop, and resize to $224\times 224$ to feed into the encoder. This gives us a TSE of shape $T\times N_f$ for some $T$. Then, for each action class, we randomly sample 50 TSEs, reshape each of them to $T_c\times N_f$ with linear interpolation, and then run 100 iterations of the DBA algorithm \cite{dba} to produce the centroid. 

We now describe our choice of the temporal sliding window widths and strides.  For \kine\ and \hmdb, given a video, \vmaev2 temporally segments the video into 5 clips of the same length and takes 3 crops at the left, center, and right to produce $5\times 3 = 15$ logit vectors, from which they then take the mean to produce the class prediction. Note that the temporal treatment is equivalent to a sliding window of width $\frac{|vid|}{5}$ and stride $\frac{|vid|}{5}$, where $|vid|$ is the video length. On the other hand, we only use the center crop, but deploy a sliding window of width $\frac{|vid|}{5}$ and stride $\frac{|vid|}{40}$, so we produce $(\frac{40}{5}\cdot (5-1)+1)\times 1=33$ logit vectors per video, with the resulting TSE having dimension $33\times N_f$. 

For \ssv2, unlike the other two datasets, \vmaev2 does not temporally segment but instead performs a strided slice on the frames with a step of 2. This means that the encoder is finetuned to an input window width of $|vid|$, which complicates our sliding window application. The vast majority of \kine\ videos are of length $\sim 300$ frames, but videos in \ssv2 vary more in frame count, ranging from the teens to over a hundred, which means its encoder window width varies more too. In order to provide \sysName\ with both constant-width and variable-width information, we apply four sliding windows with width $\{16, 32, 64, |vid|\}$ and stride 1 in parallel, and thus obtain for each video a TSE of dimension $|vid|\times (4\cdot (1408+N_c))$. The average video length in \ssv2 is $\sim 40$ frames, so on average, we produce $\frac{4\cdot 40}{33}=4.8$ times more embeddings per video than for \kine\ and \hmdb.

{\noindent\bf Applying \sysName to non-SOTA video encoders.}

Our algorithm is model-agnostic and can be applied to other encoders. To demonstrate this, we apply \sysName to other encoders that are not SOTA. The results, as presented, show a significant improvement. Here, we report as baselines numbers that the code on Hugging Face achieves on our machine rather than the numbers from the original papers, which are somewhat higher.

\begin{table}[H]
    \centering
    \begin{tabular}{|p{3cm}|p{1.2cm}|p{1.2cm}|p{1.2cm}|} \hline 
        Model&  Dataset (Clips$\times$ Crops)&  Accuracy without DejaVid& Accuracy with DejaVid\\ \hline 
        facebook/timesformer-base-finetuned-ssv2 @ huggingface \cite{facebook_timesformer_ssv2,bertasius2021space}&  SSv2\cite{ssv12} (4$|vid|\times$ 1)&  55.5\%& 57.6\%\\ \hline 
        google/vivit-b-16x2-kinetics400 @ huggingface \cite{google_vivit_kinetics400,vivit}& K400\cite{kine} (33 $\times$ 1)& 62.4\%&66.6\%\\ \hline
    \end{tabular}
\end{table}

{\noindent\bf Formulas for loss gradients $\frac{\partial L_w}{\partial U}$ and $\frac{\partial L_w}{\partial C}$.}

This section supplements \cref{subsec:backprop} by proving the differentiability of the \cref{alg:nnwdtw} neural network of \sysName, namely the detailed formulas for $\frac{\partial L_w}{\partial U}$ and $\frac{\partial L_w}{\partial C}$, which are omitted at the end of \cref{subsec:backprop}.


Recall from the end of \cref{subsec:def} that we calculate the time-weighted distance from a training or validation TSE $m$ to the centroid TSE $C_i$ of each class $i$ and then feed the class-wise distances to soft-min for class prediction. We first observe that before the soft-min, the distance calculations for each class are independent of each other; they do not share any elements of $C$ or $U$, nor do they have any inter-class connections. So we can individually calculate $\frac{\partial L_w}{\partial U[c]}$ and $\frac{\partial L_w}{\partial C[c]}$ for each class $c$, then stack them together for the final $\frac{\partial L_w}{\partial U}$ and $\frac{\partial L_w}{\partial C}$.

Note that $\frac{\partial L_w}{\partial U[c]}$ and $\frac{\partial L_w}{\partial C[c]}$ are the combination of three components: 
\[\frac{\partial L_w}{\partial U[c]}=\frac{\partial L_w}{\partial D_w[c]}\sum_l\frac{\partial D_w[c]}{\partial SC[c, l]}\frac{\partial SC[c, l]}{\partial U[c]}\]
\[\frac{\partial L_w}{\partial C[c]}=\frac{\partial L_w}{\partial D_w[c]}\sum_l\frac{\partial D_w[c]}{\partial SC[c, l]}\frac{\partial SC[c, l]}{\partial C[c]}\]

where $l$ is the index of the diagonal, $D_w\in \mathbb{R}^{N_c}$ the distance from $m$ to the centroid of each class, $SC[c,l]$ is the $l$-th skip-connection for class $c$ as in \cref{alg:nnwdtw}, and $C\in \mathbb{R}^{N_c\times T_c \times N_f}$, $U\in \mathbb{R}_{>0}^{N_c\times T_c \times N_f}$ are as defined in \cref{subsec:def}. The following tackles each of the three components respectively.

For $\frac{\partial L_w}{\partial D_w[c]}$, we use the standard cross-entropy and soft-min, so the derivative is well-known to be:
\[\frac{\partial L_w}{\partial D_w[c]} = \mathbf{y}[c] - p[c]\]
where $\mathbf{y}$ is the one-hot ground truth vector and $p[c]$ is the predicted probability of class $c$.

For $\frac{\partial D_w[c]}{\partial SC[c, l]}$, the standard trick for calculating loss gradients of a min-pooling layer is to define an indicator matrix. Note that the $l$-th min-pooling layer for class $c$ has length $\|SC[c,l]\|$. Let $R[c,l]$ of shape $\|SC[c,l]\|\times \|SC[c,l-1]\|$ be the indicator matrix of the $i$-th min-pooling for class $c$. We have:
\[
R[c,l,a,b] = 
\begin{cases} 
      1 & \text{if } b \in \{a-1, a\} \text{ and }\\
      & \text{the $a$-th output of the min-pooling}\\
      & =\text{ the $b$-th input of the min-pooling} \\
      0 & \text{otherwise}
   \end{cases}
\]

And since the min-pooling layers are chained, we have:

\[\frac{\partial D_w[c]}{\partial SC[c,l]} = \prod_{i=n+m-2}^{l+1}R[c,i]\] 

Notably, $\|SC[c, n+m-2]\|=1$, so the matrix product results in a shape of $\|SC[c, n+m-2]\|\times \|SC[c, l]\| = 1\times \|SC[c, l]\|$.

Finally, for $\frac{\partial SC[c, l]}{\partial U[c]}$, first notice that for any given $i$, $U[c, i]$ can only contribute to $SC[c,l]$ at the entry with $dist_w(U[c,i], C[c,i], m[l-i])$. Denoting $\frac{\partial SC[c, l]}{\partial U[c]}$ as $dU_l[c]\in\mathbb{R}^{\|SC[c, l]\|\times T_c \times N_f}$, we thus have:

\[
dU_l[c,i,j,f]=
\begin{cases} 
      |C[c,i+\text{start},f]-m[j,f]| \\
       \text{if $i+\text{start}+j=l$} \\
      0 \text{ otherwise}
   \end{cases}
\]

where $\text{start}=\max(0, l-\dim_0(m)+1)$ is the offset for the $0$-th element of $SC[c, l]$, as in Line 6 of \cref{alg:nnwdtw}. 

Similarly for $\frac{\partial SC[c, l]}{\partial C[c]}$, first notice that for any given $i$, $C[c, i]$ can only contribute to $SC[c,l]$ at the entry with $dist_w(U[c,i], C[c,i], m[l-i])$. Denoting $\frac{\partial SC[c, l]}{\partial C[c]}$ as $dC_l[c]\in\mathbb{R}^{\|SC[c, l]\|\times T_c \times N_f}$, we thus have:

\[
dC_l[c,i,j,f]=
\begin{cases} 
      U[c,i+\text{start},f]\cdot \text{sign}(C[c,i+\text{start},f]-m[j,f]) \\
      \text{        if $i+\text{start}+j=l$} \\
      0 \text{ otherwise}
   \end{cases}
\]

which concludes the formulas for loss gradients $\frac{\partial L_w}{\partial U}$ and $\frac{\partial L_w}{\partial C}$. This demonstrates the differentiability of the \cref{alg:nnwdtw} neural network of \sysName, which enables optimization via backpropagation.